\documentclass[a4paper,conference]{IEEEtran}
\usepackage{epsfig}
\usepackage{graphicx}
\usepackage{amsmath}
\usepackage{amssymb}
\usepackage{subfigure}
\usepackage{color}
\usepackage{epstopdf}
\usepackage{algorithm}
\usepackage{algorithmic}
\usepackage{multirow}

\usepackage{times}
\usepackage{amsmath}
\usepackage{amssymb}
\usepackage{booktabs}
\usepackage{enumerate}
\usepackage{float}
\usepackage{caption} %
\usepackage{pdfpages}

\usepackage{hyperref}
\hypersetup{
    colorlinks=true,
    linkcolor=blue,
}

\IEEEoverridecommandlockouts
\begin{document}
\title{VideoPipe 2022 Challenge: Real-World Video Understanding for Urban Pipe Inspection}

% \author{\IEEEauthorblockN{Michael Shell}
% \IEEEauthorblockA{School of Electrical and\\Computer Engineering\\
% Georgia Institute of Technology\\
% Atlanta, Georgia 30332--0250\\
% Email: http://www.michaelshell.org/contact.html}
% \and
% \IEEEauthorblockN{Homer Simpson}
% \IEEEauthorblockA{Twentieth Century Fox\\
% Email: homer@thesimpsons.com}
% \and
% \IEEEauthorblockN{James Kirk\\ and Montgomery Scott}
% \IEEEauthorblockA{Starfleet Academy\\
% San Francisco, California 96678--2391\\
% Telephone: (800) 555--1212\\



\author{\IEEEauthorblockN{
Yi Liu\IEEEauthorrefmark{1}$*$,
Xuan Zhang\IEEEauthorrefmark{1}$*$,
Ying Li\IEEEauthorrefmark{1},
Guixin Liang\IEEEauthorrefmark{2},
Yabing Jiang\IEEEauthorrefmark{2},
Lixia Qiu\IEEEauthorrefmark{2},
Haiping Tang\IEEEauthorrefmark{2},\\
Fei Xie\IEEEauthorrefmark{2},
Wei Yao\IEEEauthorrefmark{3},
Yi Dai\IEEEauthorrefmark{2}$\dag$,
Yu Qiao\IEEEauthorrefmark{1}\IEEEauthorrefmark{,4}$\dag$,
Yali Wang\IEEEauthorrefmark{1,5}$\dag$
}
%\IEEEauthorblockA{\IEEEauthorrefmark{1} Shenzhen Institutes of Advanced Technology, Chinese Academy of Sciences, China}
\IEEEauthorblockA{\IEEEauthorrefmark{1} ShenZhen Key Lab of Computer Vision and Pattern Recognition, Shenzhen Institute of Advanced Technology, \\Chinese Academy of Sciences, China}
\IEEEauthorblockA{\IEEEauthorrefmark{2} Shenzhen Bwell Technology Co., Ltd, China}
\IEEEauthorblockA{\IEEEauthorrefmark{3} Shenzhen Longhua Drainage Co., Ltd, China}
\IEEEauthorblockA{\IEEEauthorrefmark{4} Shanghai AI Laboratory, Shanghai, China}
\IEEEauthorblockA{\IEEEauthorrefmark{5} SIAT Branch, Shenzhen Institute of Artificial Intelligence and Robotics for Society}
\thanks{$*$ Yi Liu (yi.liu1@siat.ac.cn) and Xuan Zhang (xuan.zhang1@siat.ac.cn) are equally-contributed first authors. }
\thanks{
$\dag$ Yi Dai (daiyi@bominwell.com), Yu Qiao (yu.qiao@siat.ac.cn) and Yali Wang (yl.wang@siat.ac.cn) are equally-contributed corresponding authors.}
}

\DeclareRobustCommand*{\IEEEauthorrefmark}[1]{%
\raisebox{0pt}[0pt][0pt]{\textsuperscript{\footnotesize\ensuremath{#1}}}}


\maketitle

\begin{abstract}
Video understanding is an important problem in computer vision. 
Currently, 
the well-studied task in this research is human action recognition, 
where the clips are manually trimmed from the long videos, 
and a single class of human action is assumed for each clip. 
However, 
we may face more complicated scenarios in the industrial applications. 
For example, 
in the real-world urban pipe system, 
anomaly defects are fine-grained, multi-labeled, domain-relevant. 
To recognize them correctly, 
we need to understand the detailed video content.
For this reason, 
we propose to advance research areas of video understanding, 
with a shift from traditional action recognition to industrial anomaly analysis. 
In particular, 
we introduce two high-quality video benchmarks, 
namely QV-Pipe and CCTV-Pipe, 
for anomaly inspection in the real-world urban pipe systems. 
Based on these new datasets, 
we will host two competitions including (1) Video Defect Classification on QV-Pipe and (2) Temporal Defect Localization on CCTV-Pipe.
In this report, 
we describe the details of these benchmarks,
the problem definitions of competition tracks, 
the evaluation metric, 
and the result summary.
We expect that,
this competition would bring new opportunities and challenges for video understanding in smart city and beyond.
The details of our VideoPipe challenge can be found in \url{https://videopipe.github.io}.

\end{abstract}

\IEEEpeerreviewmaketitle

\section{Introduction}
\begin{figure}[t]
\begin{center}
\includegraphics[width=\linewidth]{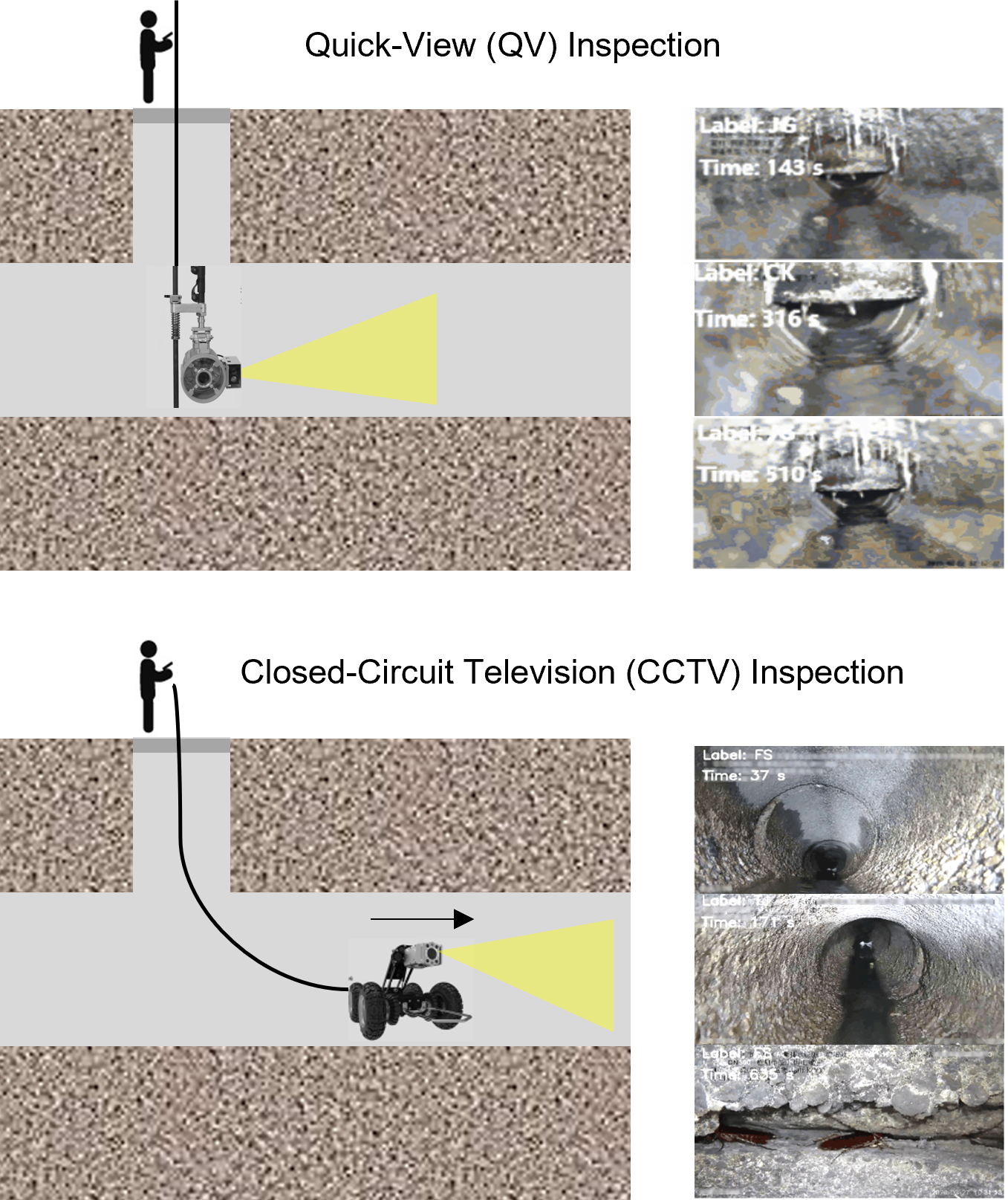}
\end{center}
%\vspace{-3mm}
\caption{Two widely-used pipe inspection methods.}
\label{fig:qv_cctv}
%\vspace{-5mm}
\end{figure}

In the last decades, 
sewer pipe system is one of the most crucial infrastructures in modern cities.
In order to ensure its normal operation, 
we need to inspect pipe defects in an effective and efficient manner.
Several technologies have been applied in the traditional pipe inspection procedure.
\cite{liu2013state} has conducted a thorough investigation and categorized them into
visual methods, 
electromagnetic methods, 
acoustic methods,
and ultrasound methods.
In particular,
Quick-View (QV) Inspection and Closed-Circuit Television (CCTV) Inspection are the most popular methods,
as shown in Figure \ref{fig:qv_cctv}.
The Quick-View (QV) Inspection is used for rapid anomaly assessment on sewer pipes, 
since the camera can only record videos on the pipe orifice.
The CCTV inspection system involves a remote-controlled robot that travels along the sewer pipe with a camera for video recording \cite{halfawy2014automated}.
Hence, 
it can get more detailed anomaly analysis for the whole pipe.
Based on these QV and CCTV videos,
the standardized protocols for manual inspection have been established and adopted in the recent years \cite{article}.
However,
it is often labor-intensive to find anomaly from hundreds of hours of videos in the complex urban pipes.

%Nowadays, most sewer inspection companies are still applying traditional manual methods, 
%which is hard and tiresome work, as the inspectors must watch the video feed for a prolonged amount of time and detect the defect of sewer pipelines.

\begin{figure*}[t]
\begin{center}
\includegraphics[width=0.95\linewidth]{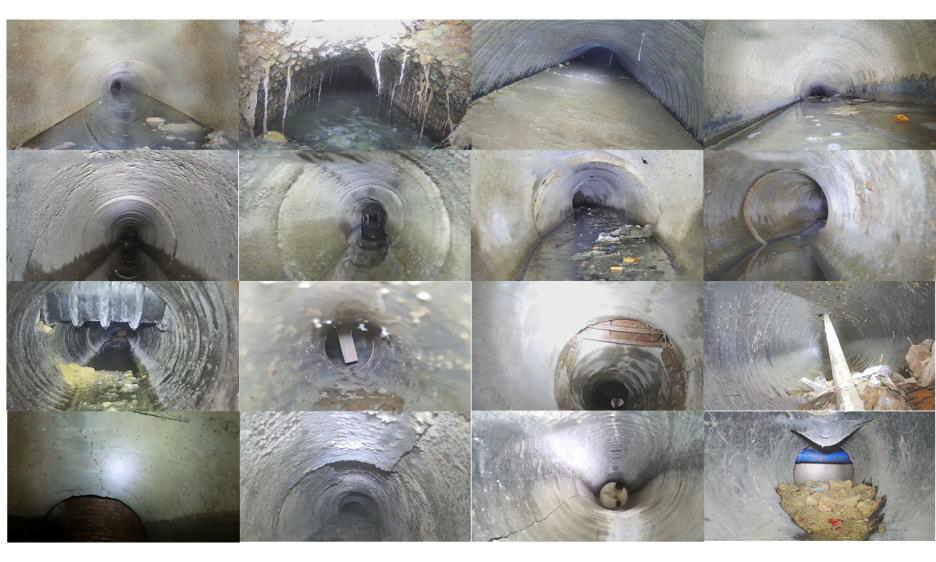}
\end{center}
% \vspace{-3mm}
\caption{Anomaly Examples in Our QV-Pipe Dataset.}
\label{fig:qv}
% \vspace{-5mm}
\end{figure*}

\begin{figure}[t]
\begin{center}
\includegraphics[width=\columnwidth]{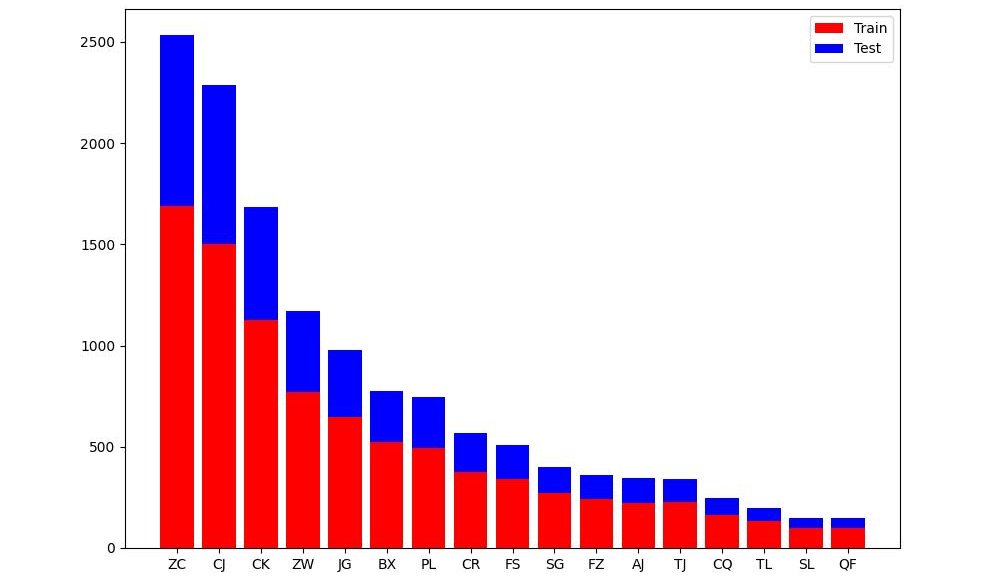}
\end{center}
% \vspace{-3mm}
\caption{Data Distribution of QV-Pipe Dataset.}
\label{fig:dis_qv}
% \vspace{-5mm}
\end{figure}

To tackle this problem,
it is essential to develop automatic inspection methods to discover sewer anomaly from large-scale pipe videos.  
Early works use hand-crafted visual features with traditional classifiers \cite{myrans2019automated,myrans2018automated,ye2019diagnosis}. 
These approaches are often limited for inspecting defects in the complex scenarios.
With fast development of deep learning frameworks,
a number of neural networks have been recently applied for 
defect classification \cite{chen2018intelligent,hassan2019underground,kumar2018automated,li2019sewer,meijer2019defect,myrans2019automated,xie2019automatic},
defect segmentation \cite{pan2020automatic,piciarelli2018vision,wang2020unified},
and
sewer defect detection \cite{fang2020sewer,moradi2017real,moradi2020automated,wang2021automated,cheng2018automated,kumar2020deep,yin2020deep}.
However,
these methods are mainly based on image-level detection.
Hence,
they are difficult to find multi-labeled and fine-grained defects,
without learning spatial-temporal contexts in the video.
Moreover,
these data sets are not available for academic research,
which blocks to develop advanced learning approaches for reliably inspecting sewer defects.

Based on the discussions above,  
we propose to establish real-world video understanding for urban pipe inspection.
To achieve this goal,
we carefully collect and annotate two new industrial video datasets, namely QV-Pipe and CCTV-Pipe, for our VideoPipe challenge.
Specifically,
QV-Pipe is used for video defect classification (Task 1) and CCTV-Pipe is used for temporal defect localization (Task 2).
In this paper,
we will introduce the details of data and tasks,
and summarize the competition results.
We expect that, 
this competition would bring new opportunities and challenges for
video understanding in smart city and beyond.

\section{Datasets and Annotation}
\begin{figure*}[t]
\begin{center}
\includegraphics[width=0.95\linewidth]{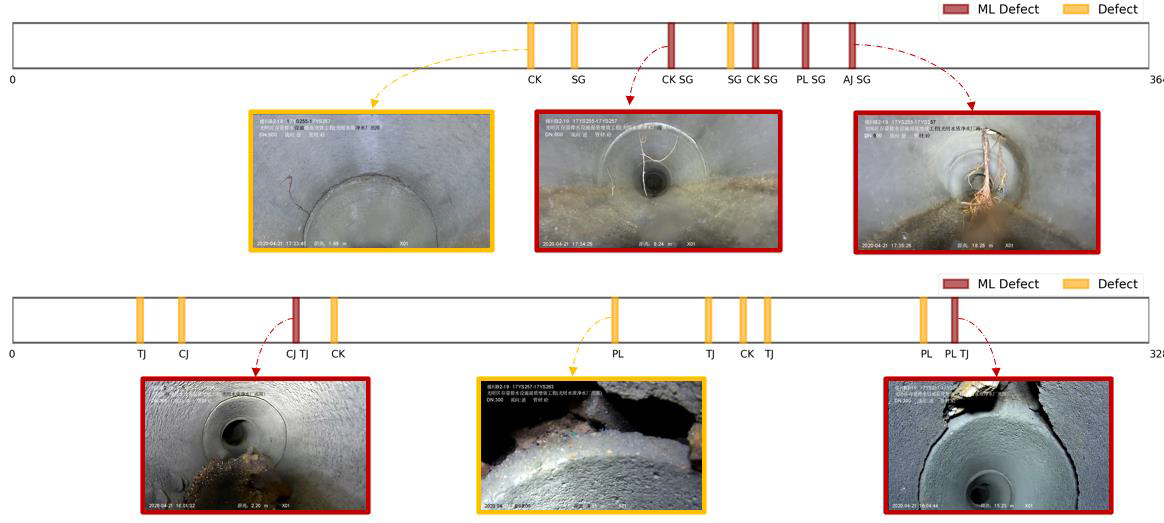}
\end{center}
% \vspace{-3mm}
\caption{Anomaly Examples of Our CCTV-Pipe Dataset. (ML: Multi-Labeled)}
\label{fig:cctv}
% \vspace{-5mm}
\end{figure*}

\begin{figure}[t]
\begin{center}
\includegraphics[width=\columnwidth]{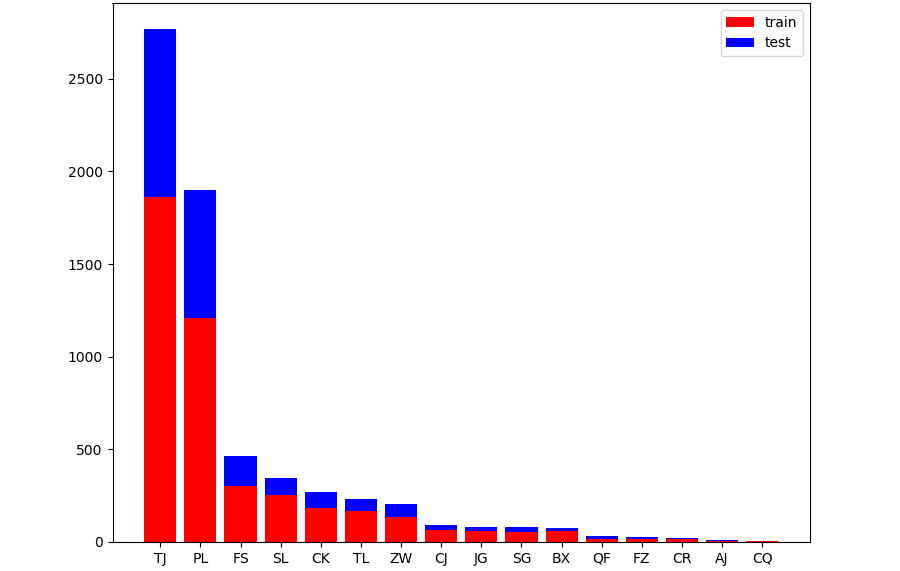}
\end{center}
% \vspace{-3mm}
\caption{Data Distribution of CCTV-Pipe.}
\label{fig:dis_cctv}
% \vspace{-5mm}
\end{figure}

\label{dataset}

In this section,
we present how these two datasets are collected and how they are constructed. 
Note that, 
%all the participants are required to sign a copyright form for academic research, 
%before getting our datasets. 
%Besides, 
these datasets are based on the real-world pipe networks. 
Hence, 
we have deleted the information of street, city and any other about privacy in our datasets.

\subsection{QV-Pipe Dataset}
\label{qvpipe}
The QV-Pipe dataset consists of 9.6k videos,
which are collected from real-world urban pipes.
The total duration of all videos exceeds 55 hours. 
Moreover,
there are 1 normal class and 16 defect classes.
Because the pipe situation is complex and multiple defects often appear at the same time, 
each video is annotated with multiple labels. 
The professional engineers are required to do this annotation procedure.
To obtain accurate annotations of defect instances, 
these engineers are asked to check all the videos multiple rounds with cross validation. 
Examples of QV-Pipe are shown in Figure \ref{fig:qv}.

The QV-Pipe video duration ranges from 0.7 seconds to 385.2 seconds. 
Each video is annotated by 1 to 5 categories. 
On average, 
each video has the duration of 20.7 seconds and 1.4 labels. 
The 9.6k videos are divided into train set and test set according to the ratio of 2:1.
As shown in Figure \ref{fig:dis_qv}, 
the data exhibits the natural long-tailed distribution.

Moreover, 
we compare it with the existing benchmarks in video anomaly detection. 
As shown in Table \ref{tab:qv-compare}, 
our QV-Pipe dataset shows the following distinct characteristics. 
First, 
compared to the existing benchmarks, 
our QV-Pipe is large scale. 
Second, 
each video in our QV-Pipe contains multiple anomaly categories, 
and these categories are fine-grained. 
Finally, 
the previous datasets mainly works on human actions. 
Alternatively, 
the domain shift is large for urban pipe inspection. 
Hence, 
our QV-Pipe brings new challenges and opportunities to understand video content for anomaly detection and beyond.

\begin{table*}[t]
\centering 
%\small
\begin{tabular}{c|ccccccc}
\hline
{\textbf{Datasets}} &{\textbf{Multi-Labeled}} &	{\textbf{Number of Classes}} &	{\textbf{Number of Videos}}	&{\textbf{Average Video Duration}} &{\textbf{Video Domain}}\\
\hline
UCSD Ped1 \cite{li2013anomaly}	& $\times$	& 2	& 70	& 5 mins & Human  Action\\
UCSD Ped2 \cite{li2013anomaly}	& $\times$	& 2	& 28	& 5 mins &Human  Action \\
Subway Entrance  \cite{adam2008robust} 	& $\times$	& 2	& 1	& 1.5 hours &Human  Action \\
Subway Exit \cite{adam2008robust} 	& $\times$	& 2	& 1	& 1.5 hours & Human  Action\\
Avenue \cite{lu2013abnormal}  	& $\times$	& 2	& 37	& 30 mins &Human  Action \\
UMN \cite{mehran2009abnormal}  	& $\times$	& 2	& 5	& 5 mins &Human  Action \\
RealWorld\cite{sultani2018real}	& $\times$	& 13	& 1,900	& 128 hours &Human  Action \\
\hline
\textbf{Our QV-Pipe} 	& \checkmark	& \textbf{17}	& \textbf{9,601}	& \textbf{55 hours} & \textbf{Pipe Defect}\\
\hline
\end{tabular}
% \vspace{3mm}
\caption{Video Anomaly Detection Benchmark Comparison. }
% \vspace{-5mm}
\label{tab:qv-compare}
\end{table*}

\begin{table*}[h]
\centering 
%\small
\begin{tabular}{c|ccccccc}
\hline
{\textbf{Datasets}} &{\textbf{Multi-Labeled}} &	{\textbf{Average Video Duration}} &	{\textbf{Types of Video Annotation}}	&{\textbf{Video Domain}}\\
\hline
THUMOS-14 \cite{idrees2017thumos}	& $\times$	& 261 s	& Instance	& Sports Action \\
ActivityNet \cite{caba2015activitynet} & 	$\times$	& 117 s	& Instance	& Daily Action \\
HACS Segment \cite{zhao2019hacs} & 	$\times$	& 149 s	& Instance	& Daily Action \\
\hline
\textbf{Our CCTV-Pipe} & 	\checkmark	& \textbf{545 s}	& \textbf{Single-frame}	& \textbf{Pipe Defect}\\
\hline
\end{tabular}
% \vspace{3mm}
\caption{Temporal Localization Benchmark Comparison. }
% \vspace{-5mm}
\label{tab:cctv_compare1}
\end{table*}

\begin{table*}[h]
\centering 
%\small
\begin{tabular}{c|ccccccc}
\hline
{\textbf{Datasets}} &{\textbf{Types of Data }} &	{\textbf{Multi-Labeled}} &	{\textbf{Number of Classes}}	&{\textbf{Number of Images/Frames}}\\
\hline
Ye et al. \cite{ye2019diagnosis}	& Image-based &	$\times$  &	7 &	1 K \\
Myrans et al.  \cite{myrans2019automated}	& Image-based &	$\times$  &	13 &	2 K \\
Chen et al. \cite{chen2018intelligent}	& Image-based &	$\times$  &	5 &	18 K \\
Li et al. \cite{li2019sewer}& Image-based &	$\times$  &	7 &	18 K \\
Kumar et al.	\cite{kumar2018automated}& Image-based &	$\times$  &	3 &	12 K \\
Meijer et al. \cite{meijer2019defect}& Image-based &	\checkmark  &	12 &	2,202 K \\
Xie et al. \cite{xie2019automatic}& Image-based &	$\times$  &	7 &	42 K \\
Hassan et al. \cite{hassan2019underground}& Image-based &	$\times$  &	6 &	24 K \\
Sewer-ML \cite{haurum2021sewer}& Image-based &	\checkmark  &	17 &	1,300 K \\
\hline
\textbf{Our CCTV-Pipe}& \textbf{Video-based} &	\checkmark  &	\textbf{16} &	\textbf{7,607 K} \\
\hline
\end{tabular}
% \vspace{3mm}
\caption{Urban Pipe Inspection Dataset Comparison. }
% \vspace{-5mm}
\label{tab:cctv_compare2}
\end{table*}

\subsection{CCTV-Pipe Dataset}
\label{cctvpipe}

Our CCTV-Pipe dataset consists of 16 defect categories including structural and functional defects in the pipe. 
It contains 575 videos with 87 hours, 
which are collected from real-world urban pipe systems. 
Different from traditional temporal action localization, 
our goal in this realistic scenario is to find preferable temporal locations of defects from a untrimmed CCTV video, instead of exact temporal boundaries. 
Hence, 
professional engineers are asked to annotate a single frame for each defect. 
The annotation procedure has been checked multiple rounds with cross validation to guarantee label quality.
We show some examples of CCTV-Pipe in Figure \ref{fig:cctv}. 
We can see that, 
several defects appear at the same temporal location. 
Additionally, 
as demonstrated in Figure \ref{fig:dis_cctv}, 
the number of defects in each category ranges from 8 to 2,770. 
Such long-tailed distribution also raises new challenges for temporal defect localization.

Moreover, 
we compare it with the existing video benchmarks in temporal localization. 
As shown in Table \ref{tab:cctv_compare1}, 
our CCTV-Pipe dataset shows the following distinct characteristics. 
First, 
compared to the existing benchmarks, 
videos in our CCTV-Pipe can be very long in practice, 
e.g., 
average video duration is 545 s.
It is quite challenging to find temporal locations of pipe defect from such long untrimmed videos. 
Second, 
instead of traditional segment annotation, 
we adopt single-frame annotation for realistic demand in urban pipe inspection. 
Moreover, 
multiple defects can densely appear at the same temporal location. 
These facts make our CCTV-Pipe as a challenging dataset for temporal localization.
Finally, 
we compare it with the existing benchmarks in pipe defect inspection. 
As shown in Table \ref{tab:cctv_compare2}, 
our dataset is based on videos, 
which is closer to urban pipe inspection in the real scenes. 
Moreover, 
our dataset is much larger than the existing ones, 
which opens new opportunities to develop powerful models for automatic defect inspection in urban pipe systems.

\section{Challenge Tasks and Evaluation}
\label{task}
Video anomaly analysis is important for urban pipe system in the real world. 
Based on our QV-Pipe and CCTV-Pipe benchmarks,
we introduce two challenging tasks in this competition,
i.e., 
Video Defect Classification and Temporal Defect Localization,
which aim at developing machine learning models to inspect pipe defects smartly and accurately.

\subsection{Task Definition}

\textbf{Task 1: Video Defect Classification}.
Quick-View (QV) Inspection is one commonly-used technology. 
However, 
it is quite labor-intensive to find defects from a huge number of QV videos. 
To tackle this problem, 
we propose a video defect classification task,
which is to predict the categories of pipe defects in a short QV video.

\textbf{Task 2: Temporal Defect Localization}.
Closed-Circuit TeleVision (CCTV) is another popular method for pipe defect inspection. 
Different from short QV videos, 
CCTV videos are much longer and record more comprehensive content in the very distant pipe. 
The main task is to discover temporal locations of pipe defects in such untrimmed videos. 
Clearly, 
manual inspection is expensive, 
based on hundreds of hours of CCTV videos. 
To fill this gap, 
we introduce this temporal localization task,
which is to find the temporal locations of pipe detects and recognizing their corresponding categories in a long CCTV video.

\subsection{Evaluation Metric}
Each task has its own evaluation metric. 
Participants have been asked to upload the results according to the specified submission format.
The submitted results have been evaluated according to different metrics.

% We will host the challenges on the platform of codalab. 
% The participant can only see the labels of training set and validation set. 
% The results on test set are submitted through the platform and evaluated by the evaluation server.

\textbf{Task1: Video Defect Classification}.
Since each video contains multiple categories, 
we use Average Precision (AP) to evaluate the recognition results on each defect category. 
Then we average AP over all the categories to obtain mAP.

\textbf{Task2: Temporal Defect Localization}.
Referring to temporal action localization, 
we use Average Precision (AP) to evaluate the defect localization results on each defect category. 
Then we average AP over all the categories to obtain mAP. 
Due to our single-frame annotations, 
we compute temporal distance between the predicted defect and the ground truth to check if this prediction is a true positive. 
Finally, 
we use the average mAP as evaluation metric, 
which is the mean of mAP with all the temporal distances (from 5 second to 15 seconds, with 5 second interval).

\begin{table*}[t]
\centering 
%\small
\begin{tabular}{c|l|c|l|l}
\hline
{\textbf{Task1}} &{\textbf{Team Name}} &{\textbf{mAP}}&{\textbf{Affiliation}} &{\textbf{Team Member}}\\
\hline
Top1 &	OverWhelmingFIt	 &	72.18 & Shanghai Paidao Intelligent Technology Co., Ltd., China & Jiawei Dong, Shuo Wang\\
Top2 &	fangxu622	     &	61.99 & Shenzhen University, China & Fang Xu\\
Top3 &	sthoduka	     &	59.91 & Hochschule Bonn-Rhein-Sieg, Germany & Santosh Thoduka\\
%4 &		huangjch526	 &	51.81 &UCAS\\
%5 &		tanghaom	 &	8.55 &\\
\hline
\end{tabular}
% \vspace{3mm}
\caption{Task 1 (Video Defect Classification): Top solutions that will be awarded. }
% \vspace{-5mm}
\label{tab:result1}
\end{table*}

\begin{table*}[t]
\centering 
%\small
\begin{tabular}{c|l|c|l|l}
\hline
{\textbf{Task2}} &{\textbf{Team Name}} & {\textbf{Avg. mAP}}&{\textbf{Affiliation}} &{\textbf{Team Member}}\\
\hline
Top1 &	OverWhelmingFIt	 & 	17.653 & Shanghai Paidao Intelligent Technology Co., Ltd., China & Jiawei Dong, Shuo Wang\\
Top2 &		sthoduka	 & 	4.325  &  Hochschule Bonn-Rhein-Sieg, Germany & Santosh Thoduka\\
\hline
\end{tabular}
% \vspace{3mm}
\caption{Task 2 (Temporal Defect Localization): Top solutions that will be awarded. }
% \vspace{-5mm}
\label{tab:result2}
\end{table*}

\subsection{Terms and Conditions}

The videos of QV-Pipe and CCTV-Pipe are authorized by Shenzhen Bwell Robotics Co.,Ltd, 
which is one key member of our organization team.
All these datasets are licensed under the Creative Commons Attribution-NonCommercial-ShareAlike 4.0 International (CC BY-NC-SA 4.0) license.
Hence, 
they must not be used for commercial purposes. 
Researchers can request access to these datasets by agreeing to terms and conditions in the following:
\begin{itemize}
\item The dataset is available for non-commercial research and educational purposes only.
\item The users agree not to reproduce, duplicate, copy, sell, trade, resell or exploit for any commercial purposes, any portion of the images, and any portion of derived data.
\item The users agree not to further copy, publish or distribute any portion of annotations of the dataset.
\item We reserve the right to terminate your access to the datasets at any time.
\end{itemize}
For our challenge,
the participants should also follow the terms and conditions below:
\begin{itemize}
\item The participants can use any pretrained models but cannot use any extra datasets for training in this competition. 
%You can use any other public dataset for pretraining and can not use any extra dataset when training the model.
\item The participants should store the submission results for evaluation purposes.
%You agree to us storing your submission results for evaluation purposes.
\item We request top-3 results to submit technical reports and/or code for us to verify submission validity.
%You agree that if you place in the top-10 at the end of the challenge you will submit your code so that we can verify that you have not cheated.
\item Any submitted reports and/or code will be used only for the sole purpose of evaluation.
\end{itemize}

\section{Organization}
% Based on the following Table \ref{tab:schedule}, w
%\textbf{
Our challenge was run according to plan. %}
First,
we built the homepage (\url{https://videopipe.github.io}) for challenge introduction and released the datasets on February 28, 2022.
Then, 
we held the challenges on the platform of CodaLab 
(Video Defect Classification: \url{https://codalab.lisn.upsaclay.fr/competitions/2232} 
and 
Temporal Defect Localization: \url{https://codalab.lisn.upsaclay.fr/competitions/2284}) 
from March 5, 2022 to May 5, 2022.
The participants could only see the labels of training set. 
The results on test set are submitted through the platform and evaluated by the evaluation server.
%We will announce the final results on May 20, 2022. 
%The awards ceremony is expected to be held on ICPR August 21, 2022.
%.

% \begin{table}[h]
% \centering 
% % \footnotesize
% \begin{tabular}{c|c}
% \hline
% {\textbf{Time}} &{\textbf{Arrangement}} \\
% \hline
% 2022-02-28	& Website and Data Release \\
% 2022-03-05	& Challenge Start \\
% 2022-05-05	& Challenge End \\
% 2022-05-15	& Winner Announcement \\
% 2022-05-20	& Challenge Report Due \\
% 2022-06-06	& Camera Ready Report Submission \\
% 2022-08-21	& Challenge Presentation Date\\
% \hline
% \end{tabular}
% % \vspace{3mm}
% \caption{Schedule of VideoPipe Challenge. }
% % \vspace{-5mm}
% \label{tab:schedule}
% \end{table}

%Some registrations come from universities in United States, United Kingdom and Germany. 
%In contrast, Task 2 has a much less valid results, mainly because of the difficulty of the defect localization.
%It is worth mentioning that the first three methods in our contest will receive prizes: for each task, the winner, second and third place will receive 1500, 500 and 250 USD.

\section{Submission and Results}

%\textbf{
Overall,
our challenge has attracted 89 participants for two tracks, %}
where
we received 51 and 38 participants respectively for Video Defect Classification and Temporal Defect Localization tracks.
%The participants come from universities, research institutions and technology companies around the world.
%Hence,
%we can draw meaningful conclusions,
%after we evaluate all the submitted results on CodaLab server.
In Table \ref{tab:result1} and Table \ref{tab:result2},
we list the top results of two tasks,
which will be awarded in the competition.
Both tasks clearly demonstrate that,
it is necessary to hold this challenge,
which encourages researchers and engineers to design more effective models for real-world video understanding. 
Note that,
according to the competition rules, 
any teams without technical report submissions will not be considered for an award.
%that does not submit a  will.
We will attach technical reports of these top results in our competition homepage,
including method framework, dataset usage, training process, and inference process.
In the following,
we briefly summarize these solutions for each task.

\textbf{Task1: Video Defect Classification}.

1. \textit{OverWhelmingFIt (Shanghai Paidao Intelligent Technology Co., Ltd., China. Team Member: Jiawei Dong, Shuo Wang). }
A comprehensive solution of frame-based method, video-based method and super-image-based method. 
In the frame-based method,
the participants use TResNet \cite{TResNet2020} (ImageNet1K pretrained) in this task,
and average prediction of all the frames as the final video prediction for inference.
In the video-based method,
the participants use Video Swin Transformer \cite{VST2021} (Swin-B backbone and Kinetics 400, Kinetics 600, or SomethingSomething V2 pretrained), 
and solve this task like traditional video classification.
In the super-image-based method,
the participants follow \cite{ICLR2022} to re-organize each video as a super-image, and use various image-based backbones (e.g., Convnext, TResNet, NFNet, EfficientNet) for classification.
Subsequently,
the participants perform weight ensemble to summarize all the models for final prediction.

2. \textit{fangxu622 (Shenzhen University, China. Team Member: Fang Xu).}
A solution with Video Swin Transformer \cite{VST2021} (Swin-B backbone).
Furthermore,
the participants use class-balanced focal loss to address class imbalance problem in the long-tailed dataset.
The implementations are mainly based on MMAction2,
with traditional settings and augmentations in video classification.

3. \textit{sthoduka (Hochschule Bonn-Rhein-Sieg, Germany. Team Member: Santosh Thoduka). }
A solution of ResNet-18 with a transformer encoder layer.
In the training phase,
5 frames are sampled randomly from a train video.
The focal loss is used to address class imbalanced problem in this task.
In the test phase,
5 frames are sampled 5 times for a test video,
and the final result is made on average of all the predictions.

\textbf{Solution Trend}: 
Overall,
the solutions of participants are promising in the track of Video Defect Classification,
based on the recent deep learning models with advanced training losses.
From aspect of models,
video backbones are often better than image backbones in this task.
Especially,
transformer-style operation can boost accuracy by learning complex long-range dependency among video frames.
The super-image solution is also interesting in this task,
which implicitly builds up frame relations via re-arranging a video as a super-image.
From aspect of losses,
the focus loss is preferable to address class imbalanced problem in such as long-tailed data distribution.

\textbf{Task2: Temporal Defect Localization}.

1. \textit{OverWhelmingFIt (Shanghai Paidao Intelligent Technology Co., Ltd., China. Team Member: Jiawei Dong, Shuo Wang). }
Basically,
the participants attempt three temporal localization solutions,
based on 
frame-level predictions, 
frame-level annotations 
and segment-level annotations.
They choose the first solution due to its better performance.
More specifically,
it is a solution of frame-level prediction by ConvNeXt.
First,
a number of pseudo-label samples are generated to increase training set,
since only single-frame annotations are available in this task.
Second,
they use ConvNeXt as image classifier for efficient training and testing.
Finally,
a number of post-processing approaches (e.g., moving average, peak finding, etc) are used to refine the final result from frame-level predictions.

2. \textit{sthoduka (Hochschule Bonn-Rhein-Sieg, Germany. Team Member: Santosh Thoduka).}
A  solution of ResNet-18 with the
addition of two linear layers is trained for the final prediction.
Due to the limited training data,
the participants enlarge the training set by frame sampling.
Specifically,
they select 15 frames in a window of $\pm$ 2.5s around each annotated frame as positive training samples, 
and randomly select extra normal frames as negative training samples. 
Moreover,
they use a validation set to determine the best threshold for each defect class.
Subsequently,
they use these thresholds to distinguish if a frame is normal or contains defects in the testing phase.

\textbf{Solution Trend}:
Compared to Video Defect Classification,
Temporal Defect Localization is much more challenging.
Since only single-frame annotations are given in this  defect moment localization task,
the participants tend to enlarge the training set,
according to frame annotations.
The post-processing step is also important to decide which frame may contain defects,
due to complex scenarios in the real-world pipe system.

%and our dataset website provides enough information to guide and inform participants.
%, respectively.
%We list the team name,  primary metric value and affiliations in the tables. 

\section{Conclusion}
This paper introduces the details of VideoPipe 2022 Challenge, 
which aims at building effective algorithms for urban pipe inspection in the real world. 
The challenge consists of two tasks including Video Defect Classification and Temporal Defect Localization,
where
we provide two large-scale video benchmarks (i.e., QV-Pipe and CCTV-Pipe) that are collected from real urban pipe systems.
The results of this competition show that,
machine learning algorithms can achieve promising results for pipe defect classification,
but still have to be promoted for temporal defect localization.
To sum up,
this competition provides new opportunities and challenges for real-world video understanding.
We expect that,
in the coming future,
these realistic video benchmarks would further show their impact on pattern recognition community and beyond.

\section{Acknowledgement}
This competition is partially supported by the National Natural Science Foundation of China (61876176,U1813218), the Joint Lab of CAS-HK, the Shenzhen Research Program (RCJC20200714114557087),  the Shanghai Committee of Science and Technology, China (Grant No. 21DZ1100100), Shenzhen Institute of Artificial Intelligence and Robotics for Society.
It is also sponsored by Shenzhen Bwell Technology Co., Ltd, China.

\bibliographystyle{IEEEtran}
\bibliography{ref}

\begin{thebibliography}{10}
\providecommand{\url}[1]{#1}
\csname url@samestyle\endcsname
\providecommand{\newblock}{\relax}
\providecommand{\bibinfo}[2]{#2}
\providecommand{\BIBentrySTDinterwordspacing}{\spaceskip=0pt\relax}
\providecommand{\BIBentryALTinterwordstretchfactor}{4}
\providecommand{\BIBentryALTinterwordspacing}{\spaceskip=\fontdimen2\font plus
\BIBentryALTinterwordstretchfactor\fontdimen3\font minus
  \fontdimen4\font\relax}
\providecommand{\BIBforeignlanguage}[2]{{%
\expandafter\ifx\csname l@#1\endcsname\relax
\typeout{** WARNING: IEEEtran.bst: No hyphenation pattern has been}%
\typeout{** loaded for the language `#1'. Using the pattern for}%
\typeout{** the default language instead.}%
\else
\language=\csname l@#1\endcsname
\fi
#2}}
\providecommand{\BIBdecl}{\relax}
\BIBdecl

\bibitem{liu2013state}
Z.~Liu and Y.~Kleiner, ``State of the art review of inspection technologies for
  condition assessment of water pipes,'' \emph{Measurement}, vol.~46, no.~1,
  pp. 1--15, 2013.

\bibitem{halfawy2014automated}
M.~R. Halfawy and J.~Hengmeechai, ``Automated defect detection in sewer closed
  circuit television images using histograms of oriented gradients and support
  vector machine,'' \emph{Automation in Construction}, vol.~38, pp. 1--13,
  2014.

\bibitem{article}
s.~Rahman and D.~Vanier, ``An evaluation of condition assessment protocols for
  sewer management,'' 01 2004.

\bibitem{myrans2019automated}
J.~Myrans, R.~Everson, and Z.~Kapelan, ``Automated detection of fault types in
  cctv sewer surveys,'' \emph{Journal of Hydroinformatics}, vol.~21, no.~1, pp.
  153--163, 2019.

\bibitem{myrans2018automated}
{Myrans, Joshua and Everson, Richard and Kapelan, Zoran}, ``Automated detection
  of faults in sewers using cctv image sequences,'' \emph{Automation in
  Construction}, vol.~95, pp. 64--71, 2018.

\bibitem{ye2019diagnosis}
X.~Ye, J.~Zuo, R.~Li, Y.~Wang, L.~Gan, Z.~Yu, and X.~Hu, ``Diagnosis of sewer
  pipe defects on image recognition of multi-features and support vector
  machine in a southern chinese city,'' \emph{Frontiers of Environmental
  Science \& Engineering}, vol.~13, no.~2, pp. 1--13, 2019.

\bibitem{chen2018intelligent}
K.~Chen, H.~Hu, C.~Chen, L.~Chen, and C.~He, ``An intelligent sewer defect
  detection method based on convolutional neural network,'' in \emph{2018 IEEE
  International Conference on Information and Automation (ICIA)}.\hskip 1em
  plus 0.5em minus 0.4em\relax IEEE, 2018, pp. 1301--1306.

\bibitem{hassan2019underground}
S.~I. Hassan, L.~M. Dang, I.~Mehmood, S.~Im, C.~Choi, J.~Kang, Y.-S. Park, and
  H.~Moon, ``Underground sewer pipe condition assessment based on convolutional
  neural networks,'' \emph{Automation in Construction}, vol. 106, p. 102849,
  2019.

\bibitem{kumar2018automated}
S.~S. Kumar, D.~M. Abraham, M.~R. Jahanshahi, T.~Iseley, and J.~Starr,
  ``Automated defect classification in sewer closed circuit television
  inspections using deep convolutional neural networks,'' \emph{Automation in
  Construction}, vol.~91, pp. 273--283, 2018.

\bibitem{li2019sewer}
D.~Li, A.~Cong, and S.~Guo, ``Sewer damage detection from imbalanced cctv
  inspection data using deep convolutional neural networks with hierarchical
  classification,'' \emph{Automation in Construction}, vol. 101, pp. 199--208,
  2019.

\bibitem{meijer2019defect}
D.~Meijer, L.~Scholten, F.~Clemens, and A.~Knobbe, ``A defect classification
  methodology for sewer image sets with convolutional neural networks,''
  \emph{Automation in Construction}, vol. 104, pp. 281--298, 2019.

\bibitem{xie2019automatic}
Q.~Xie, D.~Li, J.~Xu, Z.~Yu, and J.~Wang, ``Automatic detection and
  classification of sewer defects via hierarchical deep learning,'' \emph{IEEE
  Transactions on Automation Science and Engineering}, vol.~16, no.~4, pp.
  1836--1847, 2019.

\bibitem{pan2020automatic}
G.~Pan, Y.~Zheng, S.~Guo, and Y.~Lv, ``Automatic sewer pipe defect semantic
  segmentation based on improved u-net,'' \emph{Automation in Construction},
  vol. 119, p. 103383, 2020.

\bibitem{piciarelli2018vision}
C.~Piciarelli, D.~Avola, D.~Pannone, and G.~L. Foresti, ``A vision-based system
  for internal pipeline inspection,'' \emph{IEEE Transactions on Industrial
  Informatics}, vol.~15, no.~6, pp. 3289--3299, 2018.

\bibitem{wang2020unified}
M.~Wang and J.~C. Cheng, ``A unified convolutional neural network integrated
  with conditional random field for pipe defect segmentation,''
  \emph{Computer-Aided Civil and Infrastructure Engineering}, vol.~35, no.~2,
  pp. 162--177, 2020.

\bibitem{fang2020sewer}
X.~Fang, W.~Guo, Q.~Li, J.~Zhu, Z.~Chen, J.~Yu, B.~Zhou, and H.~Yang, ``Sewer
  pipeline fault identification using anomaly detection algorithms on video
  sequences,'' \emph{IEEE Access}, vol.~8, pp. 39\,574--39\,586, 2020.

\bibitem{moradi2017real}
S.~Moradi and T.~Zayed, ``Real-time defect detection in sewer closed circuit
  television inspection videos,'' in \emph{Pipelines 2017}, 2017, pp. 295--307.

\bibitem{moradi2020automated}
S.~Moradi, T.~Zayed, F.~Nasiri, and F.~Golkhoo, ``Automated anomaly detection
  and localization in sewer inspection videos using proportional data modeling
  and deep learning--based text recognition,'' \emph{Journal of Infrastructure
  Systems}, vol.~26, no.~3, p. 04020018, 2020.

\bibitem{wang2021automated}
M.~Wang, S.~S. Kumar, and J.~C. Cheng, ``Automated sewer pipe defect tracking
  in cctv videos based on defect detection and metric learning,''
  \emph{Automation in Construction}, vol. 121, p. 103438, 2021.

\bibitem{cheng2018automated}
J.~C. Cheng and M.~Wang, ``Automated detection of sewer pipe defects in
  closed-circuit television images using deep learning techniques,''
  \emph{Automation in Construction}, vol.~95, pp. 155--171, 2018.

\bibitem{kumar2020deep}
S.~S. Kumar, M.~Wang, D.~M. Abraham, M.~R. Jahanshahi, T.~Iseley, and J.~C.
  Cheng, ``Deep learning--based automated detection of sewer defects in cctv
  videos,'' \emph{Journal of Computing in Civil Engineering}, vol.~34, no.~1,
  p. 04019047, 2020.

\bibitem{yin2020deep}
X.~Yin, Y.~Chen, A.~Bouferguene, H.~Zaman, M.~Al-Hussein, and L.~Kurach, ``A
  deep learning-based framework for an automated defect detection system for
  sewer pipes,'' \emph{Automation in construction}, vol. 109, p. 102967, 2020.

\bibitem{li2013anomaly}
W.~Li, V.~Mahadevan, and N.~Vasconcelos, ``Anomaly detection and localization
  in crowded scenes,'' \emph{IEEE transactions on pattern analysis and machine
  intelligence}, vol.~36, no.~1, pp. 18--32, 2013.

\bibitem{adam2008robust}
A.~Adam, E.~Rivlin, I.~Shimshoni, and D.~Reinitz, ``Robust real-time unusual
  event detection using multiple fixed-location monitors,'' \emph{IEEE
  transactions on pattern analysis and machine intelligence}, vol.~30, no.~3,
  pp. 555--560, 2008.

\bibitem{lu2013abnormal}
C.~Lu, J.~Shi, and J.~Jia, ``Abnormal event detection at 150 fps in matlab,''
  in \emph{Proceedings of the IEEE international conference on computer
  vision}, 2013, pp. 2720--2727.

\bibitem{mehran2009abnormal}
R.~Mehran, A.~Oyama, and M.~Shah, ``Abnormal crowd behavior detection using
  social force model,'' in \emph{2009 IEEE conference on computer vision and
  pattern recognition}.\hskip 1em plus 0.5em minus 0.4em\relax IEEE, 2009, pp.
  935--942.

\bibitem{sultani2018real}
W.~Sultani, C.~Chen, and M.~Shah, ``Real-world anomaly detection in
  surveillance videos,'' in \emph{Proceedings of the IEEE conference on
  computer vision and pattern recognition}, 2018, pp. 6479--6488.

\bibitem{idrees2017thumos}
H.~Idrees, A.~R. Zamir, Y.-G. Jiang, A.~Gorban, I.~Laptev, R.~Sukthankar, and
  M.~Shah, ``The thumos challenge on action recognition for videos “in the
  wild”,'' \emph{Computer Vision and Image Understanding}, vol. 155, pp.
  1--23, 2017.

\bibitem{caba2015activitynet}
F.~Caba~Heilbron, V.~Escorcia, B.~Ghanem, and J.~Carlos~Niebles, ``Activitynet:
  A large-scale video benchmark for human activity understanding,'' in
  \emph{Proceedings of the ieee conference on computer vision and pattern
  recognition}, 2015, pp. 961--970.

\bibitem{zhao2019hacs}
H.~Zhao, A.~Torralba, L.~Torresani, and Z.~Yan, ``Hacs: Human action clips and
  segments dataset for recognition and temporal localization,'' in
  \emph{Proceedings of the IEEE/CVF International Conference on Computer
  Vision}, 2019, pp. 8668--8678.

\bibitem{haurum2021sewer}
J.~B. Haurum and T.~B. Moeslund, ``Sewer-ml: A multi-label sewer defect
  classification dataset and benchmark,'' in \emph{Proceedings of the IEEE/CVF
  Conference on Computer Vision and Pattern Recognition}, 2021, pp.
  13\,456--13\,467.

\bibitem{TResNet2020}
T.~Ridnik, H.~Lawen, A.~Noy, E.~B. Baruch, G.~Sharir, and I.~Friedman,
  ``Tresnet: High performance gpu-dedicated architecture,'' in
  \emph{arXiv:2003.13630}, 2020.

\bibitem{VST2021}
Z.~Liu, J.~Ning, Y.~Cao, Y.~Wei, Z.~Zhang, S.~Lin, and H.~Hu, ``Video swin
  transformer,'' in \emph{arXiv:2106.13230}, 2021.

\bibitem{ICLR2022}
Q.~Fan, C.-F.~R. Chen, and R.~Panda, ``Can an image classifier suffice for
  action recognition?'' in \emph{ICLR}, 2022.

\end{thebibliography}

%\includepdf[pages=-]{track1-top1.pdf} 
%\includepdf[pages=-]{track1-top2.pdf} 
%\includepdf[pages=-]{track1-top3.pdf} 
%\includepdf[pages=-]{track2-top1.pdf} 
%\includepdf[pages=-]{track2-top2.pdf} 

\end{document}